\documentclass{article}

\usepackage{PRIMEarxiv}

\usepackage[utf8]{inputenc}
\usepackage[T1]{fontenc}
\usepackage{hyperref}
\usepackage{url}
\usepackage{booktabs}
\usepackage{multirow}
\usepackage{amsfonts, amsmath}
\usepackage{nicefrac}
\usepackage{microtype}
\usepackage{lipsum}
\usepackage{fancyhdr}
\usepackage{graphicx}
\usepackage{xcolor}
\usepackage{enumitem}
\graphicspath{{media/}}

\title{Error-Based Bidirectional Predictive Coding\\Enables Deep,
  4-Bit Predictive Coding Networks
  \thanks{\textit{\underline{Citation}}:
  \textbf{G.~Kinman and \v{Z}.~\v{Z}ili\'c. Error-Based Bidirectional
  Predictive Coding Enables Deep, 4-Bit Predictive Coding Networks.
  Preprint, 2026.}}
}

\author{
  Garrett Kinman \\
  Department of Electrical and Computer Engineering \\
  McGill University \\
  Montr\'eal, Qu\'ebec, Canada\\
  \texttt{garrett.kinman@mail.mcgill.ca} \\
  \And
  \v Zeljko \v Zili\'c \\
  Department of Electrical and Computer Engineering \\
  McGill University \\
  Montr\'eal, Qu\'ebec, Canada\\
  \texttt{zeljko.zilic@mcgill.ca} \\
}

\begin{document}
\maketitle

\begin{abstract}
  Predictive coding networks (PCNs) promise a set of
  properties not offered by backpropagation-trained networks: learning
  from purely local information, a single network that both
  discriminates and generates, and flexible inference by energy
  minimization. However, PCNs suffer from an expensive iterative
  inference process,
  which makes hardware acceleration especially desirable.
  Thus, in pursuit of hardware-accelerated PCNs, both
  depth (to be useful) and aggressive quantization (to
  be efficient) are needed. Yet canonical state-based PC delivers neither: its
  inference signal decays exponentially with depth, so deep models
  fail to train at all, and the failure is only worsened under low
  precision. We show that a pairing of two recent methods carries the
  full PCN property set into the deep, 4-bit regime: bidirectional PC
  (bPC), which unifies discrimination and generation in one network,
  and the error-based reparameterization (ePC), in which prediction
  errors rather than states are the optimized inference variables. Error optimization delivers the output signal
  to every layer in a single backward pass, removing the depth-wise
  signal decay; a single additional element, global-norm clipping
  of the error gradients, is required to stabilize the error dynamics
  at depth. The resulting \emph{ePC+bPC} model trains stably
  for networks up to at least eight layers and retains its
  full-precision accuracy together with clean class-conditional
  generation under W4A4E4 quantization-aware training (QAT), across
  FP4 and three posit formats, where state-based bPC collapses to
  chance. On six-layer networks, where state-based bPC has collapsed
  to chance ($\sim$10\%), ePC+bPC classifies more than 50 accuracy points
  above it on only one epoch of training, even with the entire
  forward path, error tensors included,
  held at 4 bits. In addition, its class-conditional generations are
  $5$--$9\times$ better in RMSE.
  Strikingly, 4-bit quantization is nearly free for ePC+bPC:
  full-precision and 4-bit accuracies differ by at most $\sim$5
  points, and by under two in the best-performing formats, on both
  datasets. We characterize the two distinct failure modes that the
  method overcomes: bounded signal decay in state-based PC, and error
  divergence in unregularized error optimization. We then discuss the
  implications for a 4-bit-native PCN accelerator.
\end{abstract}


\section{Introduction}
\label{sec:intro}
Predictive coding (PC) is a hierarchical learning algorithm rooted in
the free-energy principle \cite{friston2005theory,friston2010free},
in which each layer of a network minimizes the discrepancy between
its own activity and a top-down prediction from the layer above
\cite{rao1999predictive}. Its appeal rests on three properties that
backpropagation (BP) does not offer:
\begin{enumerate}[leftmargin=*,topsep=4pt,itemsep=3pt]
  \item \textbf{Local learning.} Every weight update depends only on
    quantities available at adjacent layers, eliminating the global
    error-transport that dominates the routing and energy cost of BP
    on neuromorphic and edge hardware
    \cite{whittington2017approximation,salvatori2025survey}.
  \item \textbf{Native generation.} The same hierarchy that
    classifies an input simultaneously maintains a top-down model of
    it, opening the door to a single circuit that discriminates,
    generates, denoises, and completes missing data.
  \item \textbf{Flexible, energy-based inference.} Because PC infers
    by minimizing an energy over all of its variables jointly, it can
    condition on any subset of them. One circuit can therefore
    classify (clamp the input, infer the label), generate (clamp the
    label, infer the input), or complete a partially observed input,
    a flexibility that a fixed feedforward pass, running only from
    input to output, cannot offer.
\end{enumerate}
\noindent We refer to these three together as the PCN property set.

In practice, PCNs have rarely delivered this property set intact, and the
obstacle is largest where the payoff is largest: at
depth and low precision. Canonical state-based PC (sPC)
suffers an exponential decay of inference signal with depth, so under
a fixed, hardware-realistic inference budget deep models converge
slowly or not at all; and its bidirectionality has remained largely
aspirational, with discriminative variants sacrificing generation and
generative variants classifying poorly. Layered on top of this is
aggressive quantization. Four bits is not an arbitrary target. For
backpropagation-trained large language models it is bit-level
optimal: if one fixes the total memory for parameters and spends it
on either fewer high-precision weights or on more
low-precision weights, the accuracy-maximizing optimum is to use
4-bit weights. This is true across model families and scales from 19M to
176B parameters \cite{Dettmers2022TheCF}. Four bits has correspondingly become
the de facto precision for both efficient inference and,
increasingly, training
\cite{Wang2025OptimizingLL,Ashkboos2024QuaRotO4,Lee2025QrazorRA}. The
sub-8-bit datapaths that most motivate PC hardware
\cite{horowitz20141} promise the largest energy savings of
all \cite{FatemiLangroudi2018DeepLI}. For a brain-inspired algorithm,
4 bits is moreover biologically apt rather than merely expedient:
information-theoretic estimates put the storage capacity of a single
cortical synapse at $\sim$4.7 bits
\cite{Bartol2015NanoconnectomicUB,Samavat2024SynapticIS}. Yet whether
a PCN survives 4 bits at depth has, to our knowledge, never been
tested. The properties that make PCNs worth building hardware for are
thus unverified in exactly the regime that we would most like to build hardware.

Two recent results substantially address the depth and
bidirectionality obstacles. Error-based PC (ePC)
\cite{goemaere2026epcfastdeeppredictive} reparameterizes PC so that
the prediction errors, rather than the neural states, are the
variables optimized during inference. The consequence is structural.
Relaxing the states couples each variable only to its immediate
neighbours, the very locality responsible for sPC's depth-wise
signal decay. Optimizing the errors instead makes every layer a
direct function of the network output, so a single backpropagation
pass delivers the output-loss signal to all layers at once. This
eliminates sPC's exponential signal decay, and with it the depth
barrier, while provably preserving PC's equilibria and weight
gradients: the learned network is identical, only reached faster.
Bidirectional PC (bPC)
\cite{oliviers2025bidirectionalpredictivecoding} unifies the
generative and discriminative PC energies into a single objective
over a shared set of latents, so that one network genuinely does
both. Notably, the bPC authors already simulate their deep
models using ePC-style error optimization, but only as an efficiency
device for full-precision (that is, unquantized FP32) simulation, and
with only the discriminative pathway reparameterized.

In this work we take the combination of the two, which we refer to as
\textbf{ePC+bPC}, into the deep and 4-bit regime, and report
what it takes to make it work there. Our contributions are:

\begin{itemize}
  \item \textbf{The PCN property set survives the deep, 4-bit
    regime.} We show that local learning, bidirectional
    discrimination-and-generation, and energy-based inference are
    retained together at depths up to at least eight layers with the
    entire forward path, error tensors included, held at 4 bits
    (W4A4E4 QAT). This is the regime in which canonical PC fails
    outright, and the one in which a PC accelerator would most want
    to operate. Under
    the same budget, state-based bPC collapses to chance, and moving
    from full precision to 4 bits costs ePC+bPC only a few accuracy
    points. This is, to our knowledge, the first systematic study of
    bidirectional PC across depth (2--8 layers) and 4-bit number
    formats (Fig.~\ref{fig:depth}, Table~\ref{tab:main}).
  \item We identify and resolve a stability obstacle specific to the
    error-based formulation at depth: the error optimization
    diverges (error magnitudes explode by orders of magnitude)
    for networks deeper than three layers, at every precision
    including full precision. Global-norm clipping of the error
    gradients removes the divergence at all depths and formats
    (Fig.~\ref{fig:errdyn}).
  \item We characterize the two distinct failure modes the method
    overcomes: bounded signal decay in state-based PC (the network
    stalls, errors stay $O(1)$) versus error divergence in
    unregularized error optimization. This corrects a natural but
    incorrect intuition that low precision fails PCNs by driving
    errors to zero.
  \item We compare ePC+bPC across FP4~E2M1 and
    Posit$\langle$4,2$\rangle$, Posit$\langle$4,1$\rangle$,
    Posit$\langle$4,0$\rangle$, finding all four viable, and discuss
    implications for a hardware PCN accelerator with a 4-bit-native datapath.
\end{itemize}

We scope this report to multi-layer perceptrons (MLPs) on MNIST and
Fashion-MNIST. This is a deliberate choice of experimental control
rather than only a practical one. Our claim concerns the interaction
of precision, depth, and inference parameterization, and
convolutional architectures introduce a confounding factor that is
precision-independent and, in the base method, unresolved: bPC's
convolutional CIFAR-10 generations carry grid-like artifacts induced
by max-pooling in the discriminative pathway, and removing the
max-pooling to eliminate them costs roughly $23$ accuracy points
\cite{oliviers2025bidirectionalpredictivecoding}. Degradation in a
4-bit convolutional generation would therefore be difficult to
attribute to quantization rather than to that artifact. The
generative protocol we adopt is also dataset-bounded:
class-conditional generation scored against per-class mean images is
well posed for (Fashion-)MNIST, but a CIFAR-10 class mean carries
little class structure, and notably
\cite{oliviers2025bidirectionalpredictivecoding} themselves report
this protocol only on MNIST and Fashion-MNIST, switching to
reconstruction from latents for CIFAR. Holding the architecture fixed
keeps the comparison attributable to the axes we vary. Convolutional
ePC+bPC is an extension for further work, and we regard adapting PC
architectures to the convolutional setting as its own problem rather
than a free extension of this one.

\section{Background}
\label{sec:background}

\paragraph{Standard (state-based) predictive coding.}
Consider a hierarchical network with $L$ layers and activations
(``states'') $\mathbf{z}_\ell$. Each layer predicts the layer below,
$\hat{\mathbf{z}}_{\ell-1} = f(W_\ell\, \mathbf{z}_\ell)$, and the
prediction error is the direct subtraction
\begin{equation}
  \boldsymbol{\epsilon}_\ell = \mathbf{z}_{\ell-1} - \hat{\mathbf{z}}_{\ell-1}.
  \label{eq:eps_standard}
\end{equation}
The network minimizes $F =
\tfrac{1}{2}\sum_{\ell}\|\boldsymbol{\epsilon}_\ell\|^2$ by
alternating inference (relaxing the states $\mathbf{z}_\ell$
by gradient descent on $F$ with the input clamped) and
learning (updating $W_\ell$ from the converged errors). We
refer to this as state-based PC (sPC), following
\cite{goemaere2026epcfastdeeppredictive}. Crucially, in digital
simulation the inference signal in sPC decays geometrically with
distance (in layers) from the clamped output. Inference proceeds by local state
updates, so information about the clamped output spreads only one
layer per relaxation step, and every hop is attenuated: after the
first step, the signal that has reached the layer $i$ levels below the
output has been scaled by roughly $\lambda^{i}$, where the state
learning rate $\lambda<1$ is held below one for stability
\cite{goemaere2026epcfastdeeppredictive}. Supplying every layer with
usable signal therefore takes a number of relaxation steps that grows
sharply with depth, so under a fixed, small step budget the
signal never reaches the deepest layers and they receive a vanishing
learning signal. This decay is a
property of the discretized digital simulation rather than of PC
itself: \cite{goemaere2026epcfastdeeppredictive} show that with
orthogonal initialization and a large step budget sPC can still train
much deeper MLPs, but neither remedy is available cheaply on the
small-budget, low-precision hardware we target.

\paragraph{Bidirectional predictive coding.}
bPC \cite{oliviers2025bidirectionalpredictivecoding} equips each
latent with both a top-down (generative) and a bottom-up
(discriminative) prediction, sharing a single set of latents $x_l$
($x_1$ clamped to the input, $x_L$ to the label). It minimizes
\begin{equation}
  E(x, W, V) = \sum_{l=1}^{L-1}
  \frac{\alpha_{\mathrm{gen}}}{2}\big\|x_l - W_{l+1}
  f(x_{l+1})\big\|^2 + \sum_{l=2}^{L}
  \frac{\alpha_{\mathrm{disc}}}{2}\big\|x_l - V_{l-1} f(x_{l-1})\big\|^2,
  \label{eq:bpc_energy}
\end{equation}
where $W_l$ are top-down and $V_l$ bottom-up weights. The activity
update couples both errors through a bidirectional gradient
term, which is what lets one circuit both classify and generate.
Activities are initialized by a bottom-up feedforward sweep
(amortized inference).

\paragraph{Error-based predictive coding.}
ePC \cite{goemaere2026epcfastdeeppredictive} reparameterizes PC so
that the errors $\boldsymbol{\epsilon}_i$, rather than the states,
are the inference variables. The energy is unchanged, but states
become a dependent quantity recovered by a perturbed feedforward pass,
\begin{equation}
  \mathbf{s}_i := \hat{\mathbf{s}}_i + \boldsymbol{\epsilon}_i,
  \qquad \hat{\mathbf{s}}_i = f_{\theta_i}(\mathbf{s}_{i-1}).
  \label{eq:epc_states}
\end{equation}
Each inference step is a feedforward pass followed by a single
backpropagation of the output-loss gradient to every error,
$\nabla_{\boldsymbol{\epsilon}_j} E = \boldsymbol{\epsilon}_j +
(\partial \hat{\mathbf{y}}/\partial
\boldsymbol{\epsilon}_j)^{\top}\nabla_{\hat{\mathbf{y}}}\mathcal{L}$.
\cite{goemaere2026epcfastdeeppredictive} prove ePC reaches the same
equilibria and weight gradients as sPC while eliminating the
exponential signal decay: global backpropagation delivers
unattenuated signal to all layers simultaneously. Two structural
differences from Eq.~\ref{eq:eps_standard} matter below:
$\boldsymbol{\epsilon}$ is a stored, directly optimized
variable rather than a recomputed difference, and states are
reconstructed by addition (Eq.~\ref{eq:epc_states}).

\paragraph{Low-precision number formats.}
We consider four 4-bit formats. A 4-bit format can name only
$2^4=16$ distinct values, so the entire design question is
where on the real line those few values are placed. All four
formats here spend their budget in a similar way (zero, seven positive
levels, and their seven negatives) and differ primarily in the
placement (Table~\ref{tab:formats}).

FP4~E2M1 \cite{rouhani2023microscaling} is a conventional
floating-point layout: 1 sign bit, a 2-bit exponent, and a 1-bit
mantissa, giving a fixed two-values-per-octave resolution across a
modest range, $\pm[\,0.5,\,6\,]$.

Posits \cite{gustafson2017beating} instead allocate their bits
adaptively. A Posit$\langle n,es\rangle$ uses $n$ total bits
and reserves at most $es$ of them for an exponent, but its
distinguishing feature is a variable-length regime field: a
run of identical bits, terminated by the opposite bit, encoding an
integer $k$ that scales the value by $\mathit{useed}^k$ with
$\mathit{useed} = 2^{2^{es}}$. Whatever bits the sign, regime, and
exponent leave over become the mantissa. Because the regime is short
for values near $\pm 1$ and long for values near the extremes of the
range, precision is tapered: numbers close to unity get the
most mantissa bits and are represented most finely, while very large
and very small numbers are represented coarsely but do not overflow
or underflow to zero. Larger $es$ buys dynamic range and spends
near-unity precision; smaller $es$ does the reverse.

At $n=4$ this taper is severe enough to be worth reading off
directly, and the useful summary is not how many levels sit in any
one window but how widely the grid is allowed to open where the data
actually lives. Our quantizer RMS-normalizes each tensor before
snapping it to the grid (\S\ref{sec:method}), so the numbers being
quantized have unit RMS by construction and the bulk of their mass
falls in roughly $[\,0.125,\,2\,]$. A format that leaves a wide
multiplicative gap between adjacent levels inside that band must
round a large fraction of values a long way. The last column of
Table~\ref{tab:formats} reports that worst-case gap.
Posit$\langle$4,2$\rangle$ is the outlier: it spans a vast
$\pm[\,0.0039,\,256\,]$, and it pays for that range with a grid that
jumps straight from $0.125$ to $1$, an eightfold gap sitting directly
under the bulk of the distribution. FP4 and Posit$\langle$4,0$\rangle$
never open wider than twofold there. We would therefore expect
Posit$\langle$4,2$\rangle$ to suffer most, and it does
(\S\ref{sec:main_results}).

We stress that this is a statement about which format is worst, not a
full ordering. Wide range is not simply wasted: a narrow format
saturates instead, and for Posit$\langle$4,0$\rangle$, whose largest
magnitude is $4$, the saturated tail rather than the bulk is the
dominant source of quantization error on Gaussian weights. The two
effects trade off, and where they balance depends on the tensor's
distribution, which differs between weights, post-ReLU activations,
and errors. What survives across all of them is the extreme case:
a gap as wide as Posit$\langle$4,2$\rangle$'s is not recoverable.

\begin{table}[t]
  \centering
  \caption{The four 4-bit formats, given as their seven positive
    representable levels (each format also represents zero and the
    seven negatives, for 15 values in total). The last column gives
    the largest ratio between adjacent levels over
    $[\,0.125,\,2\,]$, the band holding most of the mass once tensors
    are RMS-normalized; larger means a coarser grid where the data
    sits. Posit$\langle$4,2$\rangle$ buys its wide dynamic range with
    an eightfold gap there.}
  \label{tab:formats}
  \small
  \begin{tabular}{lccc}
    \toprule
    Format & Positive levels & Range & Worst gap in $[0.125,2]$ \\
    \midrule
    FP4 E2M1 & $0.5,\ 1,\ 1.5,\ 2,\ 3,\ 4,\ 6$ & $\pm[0.5, 6]$ & $2\times$ \\
    Posit$\langle$4,0$\rangle$ & $0.25,\ 0.5,\ 0.75,\ 1,\ 1.5,\ 2,\ 4$ & $\pm[0.25, 4]$ & $2\times$ \\
    Posit$\langle$4,1$\rangle$ & $0.0625,\ 0.25,\ 0.5,\ 1,\ 2,\ 4,\ 16$ & $\pm[0.0625, 16]$ & $4\times$ \\
    Posit$\langle$4,2$\rangle$ & $0.0039,\ 0.0625,\ 0.125,\ 1,\ 2,\ 16,\ 256$ & $\pm[0.0039, 256]$ & $8\times$ \\
    \bottomrule
  \end{tabular}
\end{table}

\section{Method}
\label{sec:method}

In one sentence: we train a bidirectional PC network, but at
inference time we optimize the network's prediction errors rather
than its neural states, and we constrain the norm of the gradient
that drives that optimization. The rest of this section is those
three clauses in detail.

The network is a stack of layers carrying two sets of weights between
each neighbouring pair (Eq.~\ref{eq:bpc_energy}): top-down
(generative) weights $W$, which predict each layer from the layer
above, and bottom-up (discriminative) weights $V$, which predict each
layer from the layer below. In training we clamp the image to one end
of the stack and its label to the other, so both pathways are driven
at once. Each mini-batch is then processed in the two nested stages of
ordinary PC: an inner inference loop that settles the network's
internal variables, followed by an outer learning step that updates
the weights. Our only departure from bPC is in what the inner loop
treats as its variables:
\begin{enumerate}[leftmargin=*,topsep=2pt,itemsep=2pt]
  \item \textbf{Inference (inner loop).} We take a small, fixed number
    of gradient-descent steps ($T\!=\!8$ in our experiments) that
    adjust the stored per-layer prediction errors (rather than the
    layer states, as sPC would) so as to lower the bidirectional
    energy (Eq.~\ref{eq:bpc_energy}). Under the error
    reparameterization, only the generative (top-down) errors are
    stored and optimized. Each step therefore (i) recovers the layer
    states from those errors by a single generative (top-down)
    feed-forward pass (Eq.~\ref{eq:epc_states}), (ii) evaluates the
    discriminative (bottom-up) prediction errors directly from the
    recovered states, and (iii) backpropagates the resulting energy
    gradient to every stored error in one backward pass. There is thus
    a bottom-up computation (which forms the discriminative half of the
    energy) but no separate bottom-up relaxation sweep. It is the
    single global backward pass, rather than a layer-by-layer
    relaxation, that delivers the learning signal to all layers at once
    (\S\ref{sec:background}).
  \item \textbf{Learning (outer step).} Holding the settled errors
    fixed, we take a single Adam step on the top-down and bottom-up
    weights, using the gradients those errors define.
\end{enumerate}
One pass over the training set (a single epoch) is the entire training
run. The two baselines differ from ePC+bPC only in this inner
loop: state-based bPC relaxes the layer states instead of the errors,
and discriminative-only ePC keeps the error inference but drops the
top-down generative pathway. At test time we run only the network's
feed-forward passes, not the iterative relaxation. To classify, we
present the image alone (the label is never supplied) and read the
predicted label off a single bottom-up pass. To generate, we present a
one-hot label alone (the image is never supplied) and read the
class-conditional image off a single top-down pass.

\paragraph{Relation to prior formulations.}
Stated against the two papers we build on: ePC+bPC is bPC
(Eq.~\ref{eq:bpc_energy}) with its inference rule replaced by the
error optimization of \cite{goemaere2026epcfastdeeppredictive}. A
single set of per-layer errors reparameterizes the generative
(top-down) pathway; inference relaxes those stored errors by SGD on
the full bidirectional energy, whose discriminative (bottom-up)
term is evaluated directly from the error-recovered activities. The
error reparameterization itself is not new to deep bPC, since
\cite{oliviers2025bidirectionalpredictivecoding} already simulate
their deep models this way, but they apply it to the
discriminative pathway and only as a full-precision efficiency
device. Placing it on the generative pathway is what carries the
generative half of the property set into the deep, low-precision
regime, and it is that move, plus the gradient clipping introduced
below, that the rest of this paper evaluates.

\paragraph{Quantization-aware training.}
We apply QAT to three quantities: the weights $W$ (and $V$), the
activations $\mathbf{z}$, and the error tensors
$\boldsymbol{\epsilon}$. A quantizer $Q(\cdot)$ maps each to
one of the formats above using per-tensor RMS scaling. The forward
pass uses $Q(W),Q(\mathbf{z}),Q(\boldsymbol{\epsilon})$; backward
gradients propagate through $Q$ via a straight-through estimator
(STE). We write the joint configuration as W$b$A$b$E$b$; e.g., W4A4E4
denotes 4-bit weights, activations, and errors. Master weights remain
FP32; only the forward-path tensors are quantized. The error state is
held in $E$-format throughout inference, modelling errors that are
stored and transmitted in low precision.
Throughout, full precision, used interchangeably with FP32,
denotes the unquantized reference model: the same network, compute
budget, and training procedure with $Q(\cdot)$ replaced by the
identity, so that weights, activations, and errors are all held in
32-bit IEEE-754 floating point. Because master weights are FP32 in
every configuration, full precision differs from W$b$A$b$E$b$ only in
the forward path. It is the reference point against which our 4-bit
numbers are measured, not a claim that 32 bits are in any sense exact.

\paragraph{Gradient clipping stabilizes the error dynamics.}
The error reparameterization removes sPC's depth-wise signal decay,
but it introduces a stability hazard of its own. Because
$\boldsymbol{\epsilon}$ is directly optimized and the states are an
unbounded function of it (Eq.~\ref{eq:epc_states}), the error
optimization can diverge for deep networks: we observe the
converged error magnitude growing by several orders of magnitude with
depth (from $\sim$0.06 at $L\!=\!2$ to $\sim\!10^{3}$ at $L\!=\!6$),
at every precision including FP32 (\S\ref{sec:stability},
Fig.~\ref{fig:errdyn}). We find that global-norm clipping
\cite{pascanu2013difficulty} of the error gradients, that is,
rescaling the gradient $\boldsymbol{g}$ of the energy with respect to
the stored errors whenever its Euclidean norm exceeds a threshold $c$,
\begin{equation}
  \boldsymbol{g} \leftarrow \boldsymbol{g}\cdot\min\!\big(1,\;
  c\,/\,\|\boldsymbol{g}\|_2\big), \qquad c = 0.1,
  \label{eq:clip}
\end{equation}
applied before each error SGD step, removes the divergence at all
depths and formats, with no other change. The constraint is on the
norm of the whole gradient, so the update direction is left unchanged
and only its magnitude is bounded. This is the single
ingredient, beyond the ePC and bPC formulations themselves, that the
deep 4-bit regime requires.

\section{Experiments}
\label{sec:experiments}

\subsection{Setup}
\label{sec:setup}
We use MNIST and Fashion-MNIST (each 60k/10k split, $28\times28$
grayscale, 10 classes), run through identical pipelines. The base
architecture is an $L$-layer MLP with width-256 hidden layers and
ReLU nonlinearities, instantiated as a shared bidirectional
generator/amortizer ($x$ the one-hot label, $y$ the flattened image).
We sweep depth $L\in\{2,\dots,8\}$. Inference runs for a small,
hardware-motivated budget of $T=8$ relaxation steps per example,
applied identically to ePC+bPC and the sPC baseline (see the
dedicated discussion of the compute budget below). The error
inference learning rate is $0.05$ with global-norm clipping threshold
$c=0.1$
(Eq.~\ref{eq:clip}); the state inference learning rate for the sPC
baseline is $0.5$ (the largest value that remained stable, chosen to
give sPC the best chance at this budget); weights are trained with
Adam at $10^{-3}$ and batch size 64 for one epoch ($\approx$940
parameter updates). Each configuration is run with 5 seeds; we
report mean $\pm$ std. We report (i) test classification accuracy
from the bottom-up pass and (ii) generative quality as the RMSE
between class-conditional top-down generations and per-class mean
images (the supervised protocol of
\cite{oliviers2025bidirectionalpredictivecoding}); lower is better.

\paragraph{A fixed, equal compute budget.}
A single decision shapes every number in this paper: we hold the
compute budget fixed and identical across all methods, depths, and
precisions (one training epoch and $T\!=\!8$ inference-relaxation
steps per example) rather than tuning it per configuration. This
choice serves two ends. First, inference is expensive: every
relaxation step is a full forward pass plus a full backward pass
through the network, so the total inference cost grows with $T$. Any
practical digital accelerator will therefore want to keep $T$ small,
and the small-$T$ budget is exactly where sPC's signal decay
(\S\ref{sec:background}) bites and where a method must earn its keep.
Second, an equal budget keeps the comparison clean: ePC+bPC and the
sPC baseline then differ only in the inference parameterization, so
any gap between them is attributable to that choice and not to one
method having been handed more compute. The two inference learning
rates above are the sole per-method concession, and each was chosen to
favour its own method at this budget.

The cost of this decision is that our absolute accuracies are
budget-limited and should be read as such: they lie well short of
what longer training and more inference steps would reach for
every method (our setting is CPU-bound; \S\ref{sec:discussion}).
This shortfall is, moreover, not uniform across the sweep. A deeper
network has more layers to relax and more parameters to fit, so under
a fixed eight-step, one-epoch budget it sits further from its
own convergence than a shallow one does; deeper models would
plausibly need somewhat more inference steps and/or somewhat longer
training to close that distance. Part of ePC+bPC's gentle accuracy
decline beyond $L\!\approx\!6$ (\S\ref{sec:depth}) is therefore
expected on budget grounds alone, compounding with the weakening of
the single global-norm clipping constraint that we identify as the
other contributor (\S\ref{sec:stability}). The quantities we ask the reader to weigh
are thus the gap between methods at equal budget and the
shape of each accuracy-vs-depth curve, not the absolute
numbers or their distance from fully-trained references.

\subsection{Accuracy across network depth}
\label{sec:depth}

\begin{figure}[t]
  \centering
  \includegraphics[width=0.78\textwidth]{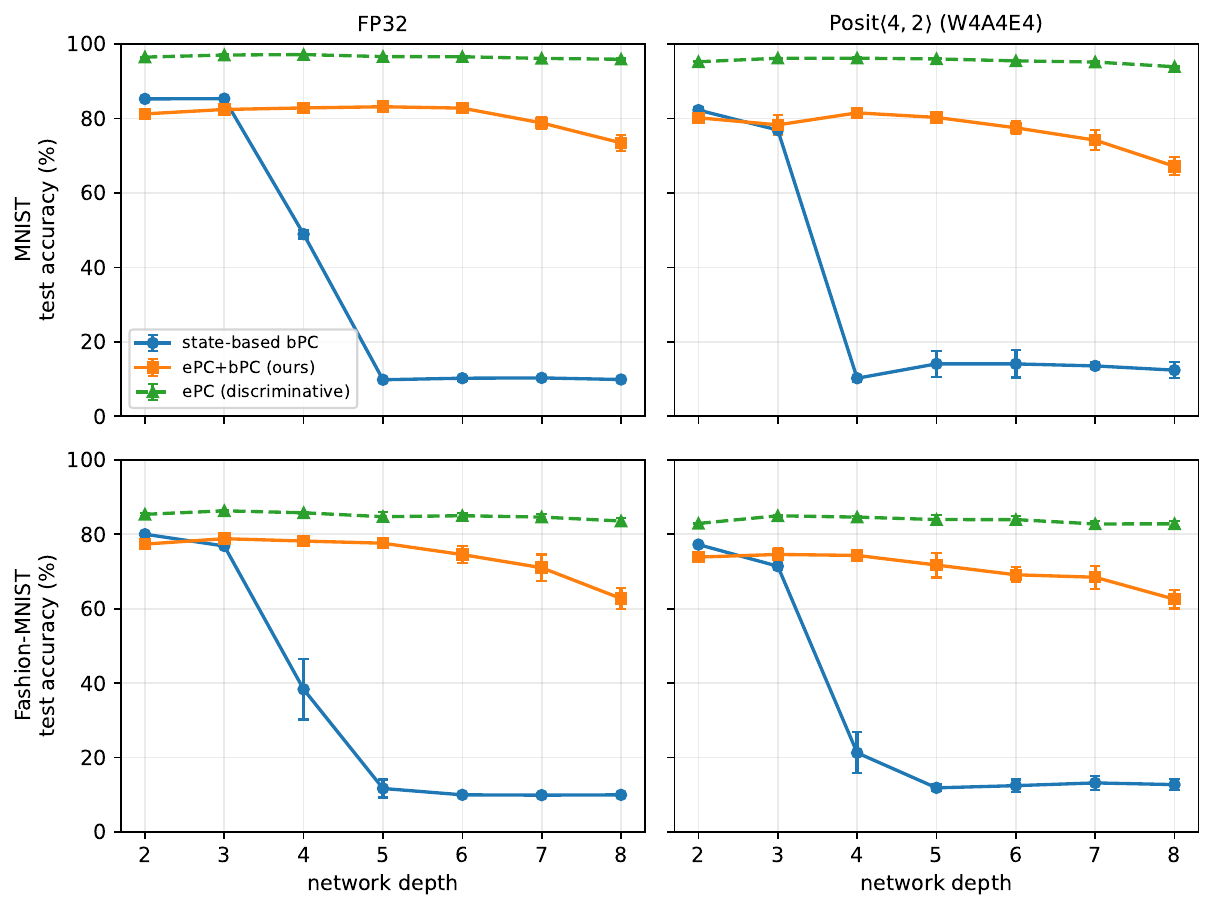}
  \caption{Test accuracy vs.\ network depth (mean $\pm$ std over 5
    seeds) on MNIST (top) and Fashion-MNIST (bottom), at full precision
    (left) and under W4A4E4 Posit$\langle$4,2$\rangle$ (right). On both
    datasets and both precisions, state-based bPC collapses to chance
    for $L\geq5$ (halving already at $L\!=\!4$) as its inference signal
    decays with depth, while ePC+bPC stays far above it at every depth,
    with only a gentle decline at the deepest layers. ePC alone
  (discriminative; dashed) is an accuracy ceiling but cannot generate.}
  \label{fig:depth}
\end{figure}

Figure~\ref{fig:depth} is the central result, and the pattern is
identical on both datasets. The 4-bit panels use
Posit$\langle$4,2$\rangle$, which is the worst-performing of
our four formats (Table~\ref{tab:main}); the curves are therefore a
conservative view of the 4-bit case. State-based bPC works
only while shallow: it reaches $80$--$85\%$ at $L\!=\!2$--$3$,
halves by $L\!=\!4$ (e.g.\ $49\%$ MNIST / $38\%$ Fashion at FP32),
and sits at chance ($\sim$10\%) for all $L\!\geq\!5$. ePC+bPC
stays far above this collapse at every depth: on MNIST it holds
$81$--$83\%$ (FP32) and $77$--$82\%$ (Posit$\langle$4,2$\rangle$)
through $L\!=\!6$, declining gently to $73\%$/$67\%$ by $L\!=\!8$; on
Fashion-MNIST it holds $75$--$79\%$ and $69$--$75\%$ through
$L\!=\!6$, declining to $\sim$63\% by $L\!=\!8$. Even at the deepest
layers the two methods differ by $60$+ accuracy points; the gentle
decline beyond $L\!\approx\!6$ reflects our fixed compute budget
biting harder on deeper networks (\S\ref{sec:setup}) together with the
point at which a single global-norm clipping constraint begins to
lose control of the error dynamics (\S\ref{sec:stability}), not a
return to the sPC failure mode. The
discriminative-only ePC ceiling ($\sim$96\% MNIST, $\sim$85\%
Fashion) shows the cost of bidirectionality: ePC+bPC trades raw
accuracy for the generative pathway, the property that makes it more
than a classifier.

The sPC collapse here is specific to the small-budget regime, and we
want to be precise about what it does and does not show. It is
not a claim that sPC cannot represent deep functions: given
orthogonal initialization and a much larger relaxation budget, sPC
trains MLPs of $20$ layers on these same datasets
\cite{goemaere2026epcfastdeeppredictive}. What Figure~\ref{fig:depth}
shows is that at the fixed eight-step budget a low-power accelerator
may reasonably be able to afford, sPC's signal decay leaves deep
layers untrained, whereas
error optimization delivers the output signal to every layer in a
single pass and so stays trainable across all depths we tested. The
relevant axis for hardware is the inference-step budget, and along
that axis the gap is decisive.

\subsection{Main results: 4-bit is nearly free, on both datasets}
\label{sec:main_results}

Table~\ref{tab:main} reports the deep regime ($L\!=\!6$) across all
formats for MNIST and Fashion-MNIST. ePC+bPC retains its accuracy and
a tight generative RMSE ($\sim$0.07--0.10) at every 4-bit
format, within $1$--$5$ points of its own FP32 number; the entire
forward path, including the error tensors, runs at 4 bits at almost
no cost. State-based bPC is at chance throughout, with generative
RMSE $5$--$9\times$ worse.

\begin{figure}[t]
  \centering
  \includegraphics[width=0.95\textwidth]{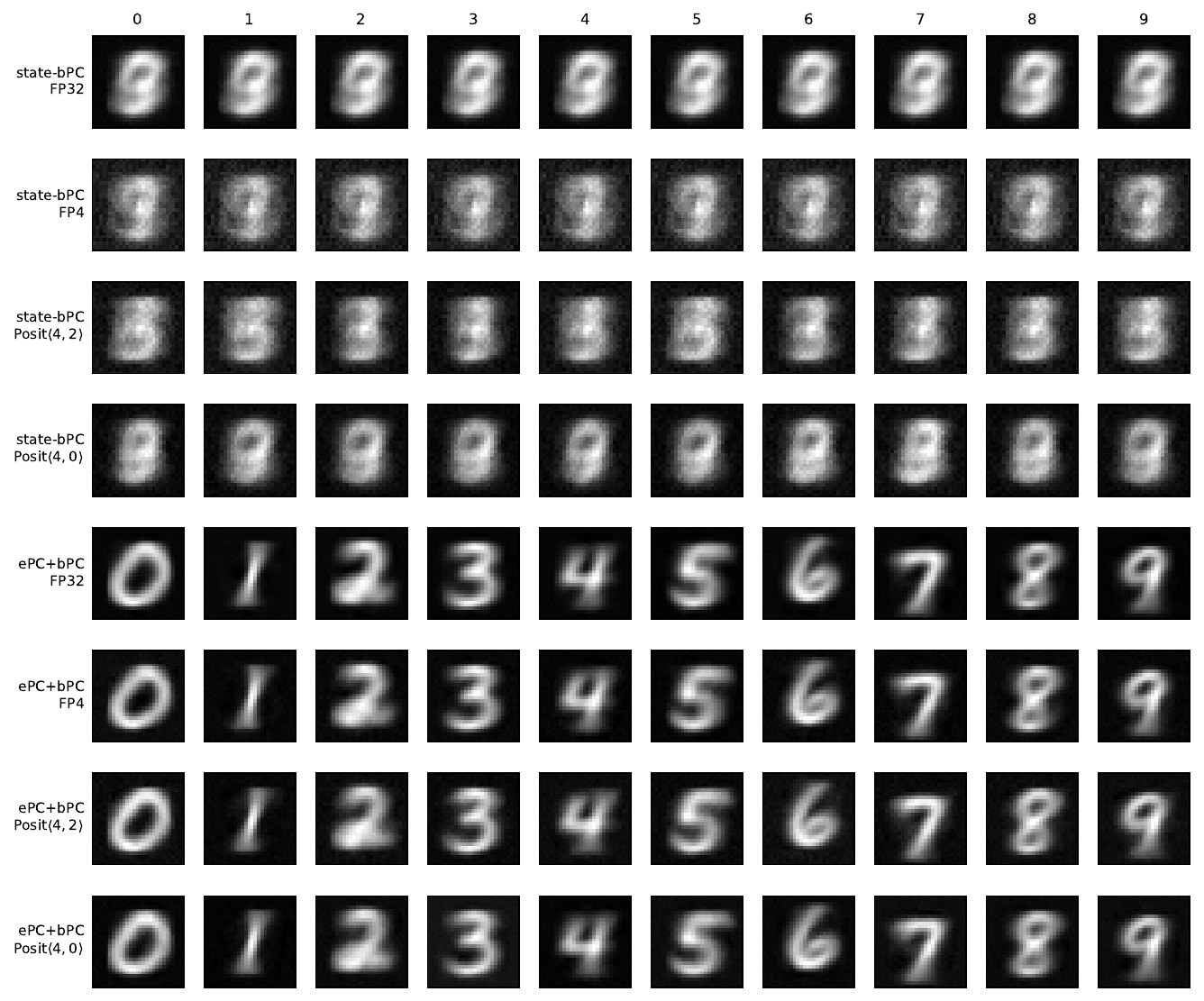}
  \caption{Class-conditional MNIST generations at depth $L\!=\!6$.
    Top four rows: state-based bPC; bottom four rows: ePC+bPC; one row
    per number format. ePC+bPC generates clean digits at FP32 and all
    4-bit formats; state-based bPC, unable to train at this depth,
  produces label-independent noise.}
  \label{fig:samples_mnist}
\end{figure}

\begin{figure}[t]
  \centering
  \includegraphics[width=0.95\textwidth]{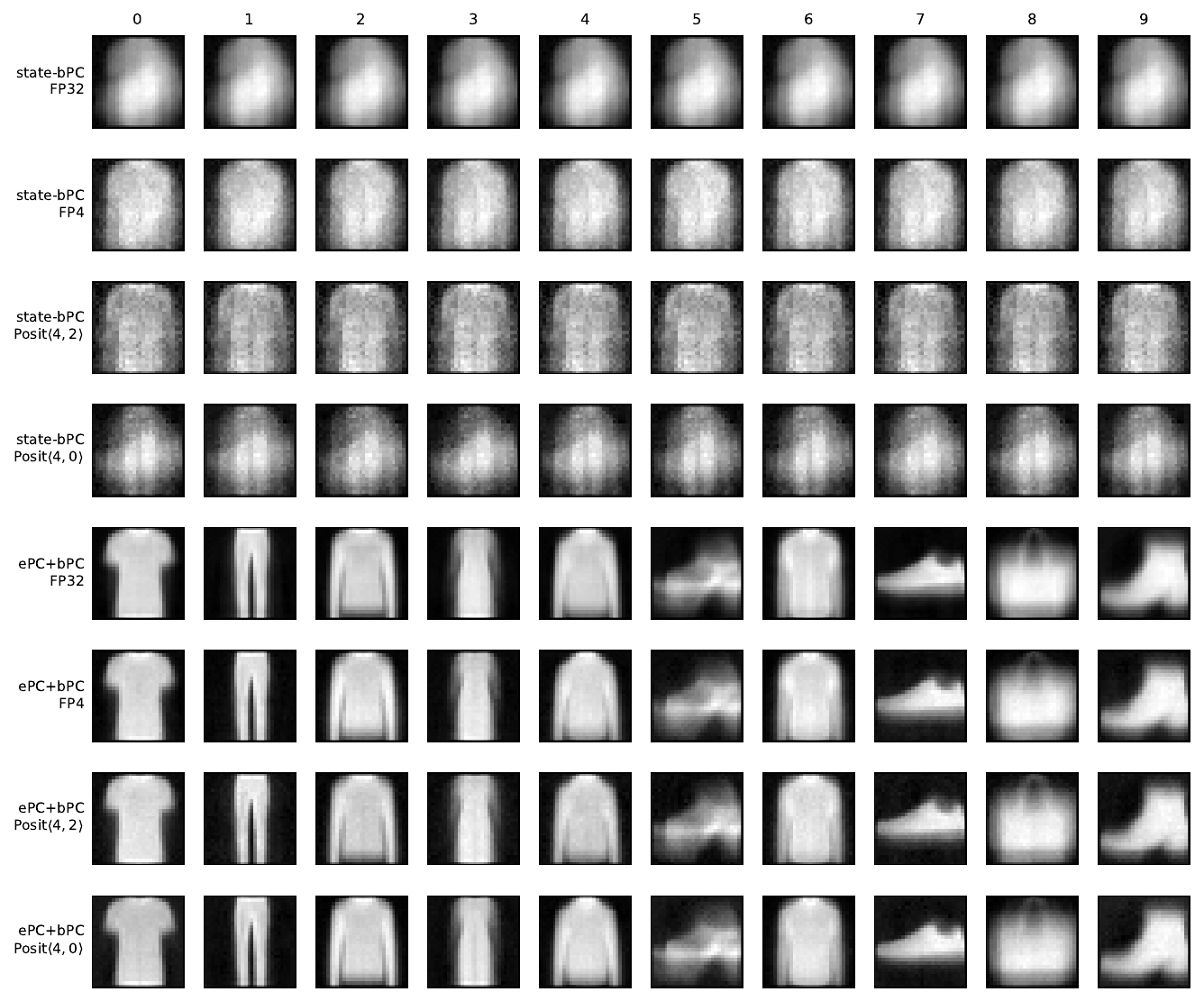}
  \caption{Class-conditional Fashion-MNIST generations at depth
    $L\!=\!6$ (same layout as Fig.~\ref{fig:samples_mnist}). ePC+bPC
    renders recognizable garments across all precisions; state-based
  bPC collapses to noise.}
  \label{fig:samples_fashion}
\end{figure}

The one format that separates from the pack does so in the direction
Table~\ref{tab:formats} predicts. Posit$\langle$4,2$\rangle$ is the
weakest 4-bit format on both datasets ($-5.3$ and $-5.4$
points relative to FP32), and it is also the format whose grid opens
widest under the bulk of the data, jumping eightfold from $0.125$ to
$1$ where the others never exceed a factor of four. The other three
land within $0.4$--$3.2$ points of FP32. We do not read a finer
ranking into those three: they lie within a couple of points of one
another and do not order consistently across the two datasets
(Posit$\langle$4,0$\rangle$ is best on MNIST,
Posit$\langle$4,1$\rangle$ on Fashion-MNIST). That instability is to
be expected rather than noise alone, since the relative merit of the
remaining formats depends on the distribution of the tensor being
quantized, and weights, post-ReLU activations, and error tensors do
not share one. The usable conclusion for a hardware designer is a
floor rather than an ordering: avoid leaving a wide gap under the
bulk of the distribution, and any of these formats will serve.
Figures~\ref{fig:samples_mnist}--\ref{fig:samples_fashion} show the
corresponding generations: ePC+bPC produces clean, class-distinct
images at all precisions, while state-based bPC produces
label-independent noise.

\begin{table}[t]
  \centering
  \caption{Test accuracy (\%) and generative RMSE (lower better) at
    depth $L\!=\!6$, mean $\pm$ std over 5 seeds. ePC (discriminative)
    has no generative pathway and was run only at FP32 and
  Posit$\langle$4,2$\rangle$.}
  \label{tab:main}
  \small
  \begin{tabular}{llccccc}
    \toprule
    & Method & FP32 & FP4 E2M1 & Posit$\langle$4,2$\rangle$ &
    Posit$\langle$4,1$\rangle$ & Posit$\langle$4,0$\rangle$ \\
    & & Acc / RMSE & Acc / RMSE & Acc / RMSE & Acc / RMSE & Acc / RMSE \\
    \midrule
    \multirow{3}{*}{\rotatebox{90}{MNIST}}
    & bPC, state-based   & $10.2$ / $0.45$ & $12.0$ / $0.58$ & $14.1$
    / $0.50$ & $13.5$ / $0.50$ & $11.2$ / $0.50$ \\
    & ePC (no bidirect.) & $96.6$ / ---   & ---             & $95.5$
    / ---    & ---             & ---            \\
    & \textbf{ePC+bPC (ours)} & $\mathbf{82.8}$ / $\mathbf{0.09}$ &
    $\mathbf{79.6}$ / $\mathbf{0.08}$ & $\mathbf{77.5}$ /
    $\mathbf{0.09}$ & $\mathbf{80.1}$ / $\mathbf{0.10}$ &
    $\mathbf{82.4}$ / $\mathbf{0.09}$ \\
    \midrule
    \multirow{3}{*}{\rotatebox{90}{\small Fashion}}
    & bPC, state-based   & $10.0$ / $0.59$ & $13.6$ / $0.66$ & $12.5$
    / $0.64$ & $13.0$ / $0.61$ & $12.3$ / $0.64$ \\
    & ePC (no bidirect.) & $85.0$ / ---   & ---             & $83.9$
    / ---    & ---             & ---            \\
    & \textbf{ePC+bPC (ours)} & $\mathbf{74.5}$ / $\mathbf{0.09}$ &
    $\mathbf{73.0}$ / $\mathbf{0.07}$ & $\mathbf{69.1}$ /
    $\mathbf{0.09}$ & $\mathbf{73.3}$ / $\mathbf{0.08}$ &
    $\mathbf{72.5}$ / $\mathbf{0.08}$ \\
    \bottomrule
  \end{tabular}
\end{table}

\subsection{Error dynamics: decay, divergence, and gradient clipping}
\label{sec:stability}

\begin{figure}[t]
  \centering
  \includegraphics[width=0.55\textwidth]{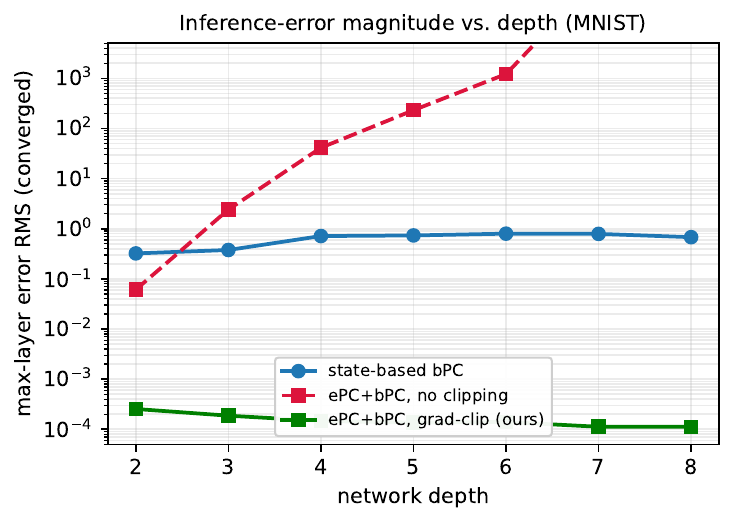}
  \caption{Converged max-layer inference-error magnitude vs.\ depth
    (averaged over formats and seeds). State-based bPC errors stay
    bounded ($O(1)$); it fails by signal decay, not blow-up.
    Unregularized ePC+bPC errors diverge, growing from
    $\sim$0.06 at $L\!=\!2$ to $\sim\!10^{3}$ at $L\!=\!6$ and
    $\sim\!10^{5}$ by $L\!=\!8$. Global-norm clipping (ours) holds them
  at $\sim\!10^{-4}$ across all depths.}
  \label{fig:errdyn}
\end{figure}

The two methods fail for opposite reasons (Fig.~\ref{fig:errdyn}).
State-based bPC fails by bounded decay: its converged
error magnitudes stay $O(1)$ at every depth, but the inference signal
that reaches deep layers shrinks geometrically, so the network simply
stalls. Unregularized ePC+bPC fails by divergence:
because the errors are directly optimized and the states depend on
them without bound (Eq.~\ref{eq:epc_states}), the converged error
magnitude explodes with depth (from $\sim$0.06 at $L\!=\!2$ to
$\sim\!10^{3}$ at $L\!=\!6$ and $\sim\!10^{5}$ by $L\!=\!8$), and
this happens at FP32 as well as at 4 bits. Notably, this divergence
appears tied to the bidirectional coupling rather than to error
optimization in general: purely discriminative ePC trains $20$-layer
MLPs on these datasets without it
\cite{goemaere2026epcfastdeeppredictive}, so it is the addition of
the top-down generative error that destabilizes the iteration at
depth. Gradient clipping (Eq.~\ref{eq:clip}) holds the error
magnitude at $\sim\!10^{-4}$ across all depths, recovering stable
training everywhere.

We stress that neither failure is the naively-assumed ``errors
quantize to zero.'' Low precision is not the proximate cause of
either collapse: state-based PC fails by depth-wise decay independent
of precision (Fig.~\ref{fig:depth}, left), and the error divergence
is a property of the unregularized optimization, not of the grid.
This corrects a tempting catastrophic-cancellation intuition and
reveals the real obstacles in (i) inference-signal decay and (ii)
error-optimization stability, both of which ePC, with clipping, resolves.

\subsection{Ablation}
\label{sec:ablations}
Table~\ref{tab:ablation} isolates each ingredient at $L\!=\!6$ under
Posit$\langle$4,2$\rangle$. Removing the error reparameterization
(state-based bPC) collapses accuracy and generation. Removing
bidirectionality (ePC alone) recovers high accuracy but forfeits
generation. Removing gradient clipping from ePC+bPC reintroduces the
divergence of \S\ref{sec:stability}. Only the full method is both
accurate and generative at depth and 4 bits. Fashion-MNIST replicates
the pattern exactly (state-based $12.5\%$, ePC-alone $83.9\%$,
ePC+bPC $69.1\%$ at the same setting).

\begin{table}[t]
  \centering
  \caption{Ablation at $L\!=\!6$, W4A4E4 Posit$\langle$4,2$\rangle$
  (mean over 5 seeds).}
  \label{tab:ablation}
  \begin{tabular}{lcc}
    \toprule
    Configuration & Test Acc (\%) & Generation \\
    \midrule
    state-based bPC (no ePC)        & $14.1$ & poor (RMSE 0.50) \\
    ePC alone (no bidirectionality) & $95.5$ & --- (none) \\
    ePC+bPC, no gradient clipping   & $13.5$ & diverged \\
    \textbf{ePC+bPC (full)}         & $\mathbf{77.5}$ & good (RMSE 0.09) \\
    \bottomrule
  \end{tabular}
\end{table}

\section{Discussion}
\label{sec:discussion}

\subsection{Why ePC+bPC reaches the deep, 4-bit regime}
Two mechanisms are at work. First, depth: error optimization
delivers the output-loss gradient to every layer in one backward pass
(Eq.~\ref{eq:epc_states}), eliminating the geometric signal decay
that confines sPC to shallow networks; this is what flattens the
accuracy-vs-depth curve (Fig.~\ref{fig:depth}). Second,
stability: the same direct parameterization that removes decay
makes the error iteration prone to divergence at depth, which
global-norm clipping (Eq.~\ref{eq:clip}) controls
(Fig.~\ref{fig:errdyn}). Given both, low precision is remarkably
benign: per-tensor-scaled 4-bit weights, activations, and errors cost
at most $\sim$5 accuracy points relative to FP32, and under two
points in the best-performing formats (Table~\ref{tab:main}), because
the quantities being quantized are well-conditioned once the dynamics
are stable. That 4 bits is nearly free here is the predictive-coding
analogue of a finding established for backpropagation-trained LLMs,
where 4-bit is the bit-level-optimal precision for a fixed memory
budget and accuracy degrades sharply only below it
\cite{Dettmers2022TheCF}. Our result extends that design point from
post-training weight quantization of a feedforward transformer to
quantization-aware training of a local, bidirectional
algorithm whose entire forward path (weights, activations, and the
error tensors) is held at 4 bits. Whether
PCNs exhibit an analogous cliff below 4 bits is a natural question we
leave to future work.

\subsection{Implications for hardware co-design}
\begin{itemize}[leftmargin=*,topsep=2pt,itemsep=2pt]
  \item \textbf{A 4-bit-only datapath is viable for deep PC.} W4A4E4
    ePC+bPC matches its FP32 self to within $\sim$5\% at six layers,
    demonstrating that the entire forward path, including error
    tensors, can be 4 bits, substantially reducing register/SRAM area
    relative to an 8-bit baseline.
  \item \textbf{Do not leave a wide gap under the bulk of the
    distribution.} All four formats are usable, and the constraint we
    can state with confidence is a floor rather than a ranking.
    Because per-tensor RMS scaling gives every quantized tensor unit
    RMS, a format is penalised in proportion to how far apart its
    levels sit where that mass lands. The worst format on both
    datasets is exactly the one that opens widest there
    (Posit$\langle$4,2$\rangle$, an eightfold gap from $0.125$ to $1$;
    Tables~\ref{tab:formats} and \ref{tab:main}), while the formats
    holding that gap to a factor of four or less are all within a few
    points of FP32. We would not push this further into a
    recommendation among those three: extra dynamic range is not
    simply wasted, since a narrow format saturates the tail instead,
    and the balance between coarse-grid error and saturation error
    shifts with the tensor's distribution. Low-$es$ posits are separately
    attractive on area grounds: at $n=4$ the regime decoder is small,
    and Posit$\langle$4,0$\rangle$ needs no exponent field at all.
  \item \textbf{Error optimization adds a backward pass per inference
    step.} Each step needs a full backpropagation pass through the
    network on top of the forward pass, whereas state-based PC updates
    its states with purely local operations. This is what trades away
    strict neuromorphic locality, but it is also what lets the output
    signal reach every layer at once, permitting both depth and a
    4-bit datapath. For a digital datapath this is the right trade,
    and the equivalence proof of
    \cite{goemaere2026epcfastdeeppredictive} guarantees the learned
    weights are unchanged.
\end{itemize}

\subsection{Limitations and future work}
The bidirectional model classifies below a pure discriminative ePC
(e.g.\ $\sim$80\% vs.\ $\sim$96\% on MNIST); its value is the
combination of competitive accuracy with generation at 4 bits,
not raw accuracy. Absolute accuracies are further held down by our
one-epoch training budget (\S\ref{sec:setup}); we expect longer
training to lift all methods, but report the equal-budget comparison
that the relative claims rest on. Beyond $L\!\approx\!6$, ePC+bPC
accuracy declines (MNIST FP32 $83\%\!\to\!73\%$ from $L\!=\!6$ to
$L\!=\!8$, with a steeper drop on Fashion-MNIST), indicating that the
stabilizing effect of a single global-norm clipping constraint
weakens at extreme depth even as it prevents outright divergence. Our experiments are also
CPU-bound, which capped the practical sweep scale (seeds, depth, and
especially CIFAR-scale inputs); this is a limitation of the present
study rather than of the method, and it underscores the work's
motivation: the iterative energy-minimization inference at the
core of PC is poorly matched to general-purpose processors, and a
native low-precision PCN accelerator is precisely what would make
deep PCNs practical at
scale. Three scope caveats bear emphasis. First, ``deep'' here means up
to eight layers; we have not yet probed the $20$+ layer regime in
which ePC's advantage over sPC is largest
\cite{goemaere2026epcfastdeeppredictive} and in which, as the
$L\!=\!7$--$8$ decline already hints, the bidirectional error
dynamics will demand more than a single global-norm clipping
constraint. Second, our
comparison fixes a small ($T\!=\!8$) inference budget and standard
initialization; we do not claim sPC is fundamentally incapable at
depth (with orthogonal initialization and a far larger step budget
it can train deeper networks \cite{goemaere2026epcfastdeeppredictive});
we claim only that error optimization reaches the same regime without
either expense, which is what matters on hardware. Third, our
conclusions are established for fully-connected architectures. As
argued in \S\ref{sec:intro}, this holds the architecture fixed so
that the precision and depth axes remain attributable, and it avoids
a generative artefact that is unresolved in convolutional bPC at full
precision \cite{oliviers2025bidirectionalpredictivecoding}; the cost
is that we do not yet know whether 4 bits is equally cheap once a
convolutional generative pathway is in play. Convolutional ePC+bPC on
CIFAR-10, scaling to the deeper regime, a closer study of the
error-clipping threshold, and an end-to-end hardware accelerator at
W4A4E4 are the immediate next steps.

\section{Conclusion}
\label{sec:conclusion}
The appeal of predictive coding has always been a set of
properties (local learning, a single circuit that discriminates and
generates, flexible energy-based inference) that matter most on
exactly the deep, low-precision regime where canonical PC fails. We
have shown that this property set survives into that regime by pairing
bidirectional PC with the error-based reparameterization, and
stabilizing the error dynamics by clipping the norm of the error
gradients. Combined,
these methods yield networks that train to at least eight layers and
retain their full-precision accuracy with clean generation under
W4A4E4 QAT (across FP4 and three posit formats, on both MNIST and
Fashion-MNIST), where state-based bPC collapses to noise, and at
almost no cost relative to full precision. We further clarified the
mechanism: the obstacles are inference-signal decay and
error-optimization stability, not the often-assumed quantization of
errors to zero. The result clears a practical path to a deep,
4-bit-native datapath for PCN accelerators.

\bibliographystyle{unsrt}
\bibliography{references}

\end{document}